%% file: neurips_2026.tex
\documentclass{article}

\usepackage[preprint]{neurips_2026}

\usepackage[utf8]{inputenc} 
\usepackage[T1]{fontenc}    
\usepackage{hyperref}       
\usepackage{url}            
\usepackage{booktabs}       
\usepackage{amsfonts}       
\usepackage{nicefrac}       
\usepackage{microtype}      
\usepackage{xcolor}         

\usepackage{graphicx}
\usepackage{subcaption}
\usepackage{enumitem}       
\usepackage{float}          
\usepackage[mathcal]{euscript}

\title{Cluster Attention for Graph Machine Learning}

\author{%
  Oleg Platonov\thanks{Corresponding author} \\
  HSE University, Yandex Research \\
  \texttt{olegplatonov@yandex-team.ru} \\
  \And
  Liudmila Prokhorenkova \\
  Yandex Research \\
  \texttt{ostroumova-la@yandex-team.ru} \\
}

\begin{document}

\maketitle

\begin{abstract}
  Message Passing Neural Networks have recently become the most popular approach to graph machine learning tasks; however, their receptive field is limited by the number of message passing layers. To increase the receptive field, Graph Transformers with global attention have been proposed; however, global attention does not take into account the graph topology and thus lacks graph-structure-based inductive biases, which are typically very important for graph machine learning tasks. In this work, we propose an alternative approach: cluster attention (CLATT). We divide graph nodes into clusters with off-the-shelf graph community detection algorithms and let each node attend to all other nodes in each cluster. CLATT provides large receptive fields while still having strong graph-structure-based inductive biases. We show that augmenting Message Passing Neural Networks or Graph Transformers with CLATT significantly improves their performance on a wide range of graph datasets including datasets from the recently introduced GraphLand benchmark representing real-world applications of graph machine learning.
\end{abstract}

\section{Introduction}

Graphs are a natural way to represent data from different domains such as social networks (both real-life and virtual), computer networks, transportation networks, co-purchasing networks, molecules, connectomes, neural network architectures, various physical and biological systems, or even interconnected abstract concepts. Thus, machine learning on graph-structured data, henceforth referred to as Graph Machine Learning (GML), has attracted a lot of attention recently. In particular, Graph Neural Networks (GNNs) \citep{duvenaud2015convolutional, kipf2017semi, gilmer2017neural, hamilton2017inductive} have become the most popular models for most GML tasks in the past decade. Most modern GNNs can be unified by the Message Passing Neural Networks (MPNNs) framework \citep{gilmer2017neural}. In this framework, within each neural network layer each node in the graph sends messages to its neighbors and aggregates incoming messages to form its new representation based on them. The graph-structure based relational inductive biases of MPNNs, which send messages along graph edges, turned out to be very effective for modeling many real-world networks \citep{battaglia2018relational}. However, in each MPNN layer, the model only exchanges information between neighboring nodes. Thus, the model's receptive field, i.e., the maximum shortest path distance information can travel between nodes in the graph, is equal to the number of message passing layers. This has been argued to be a limitation of MPNNs as it prevents taking into account long-range interaction between nodes that might happen in some networks. Thus, Graph Transformers with global (all-to-all) attention have been proposed as an alternative neural model architecture for GML \citep{kreuzer2021rethinking, ying2021transformers, wu2022nodeformer, rampavsek2022recipe}, inspired by the success of the Transformer architecture in the field of natural language processing \citep{vaswani2017attention}. However, all-to-all attention does not have any information about the graph structure, which is arguably very important for solving GML tasks, and thus this information has to be integrated into the model in some other ways (typically via positional encodings or MPNN-Transformer hybridization).

It has been observed that many real-world networks exhibit community structure~--- presence of well-defined clusters of densely connected nodes with sparser connections between these clusters \citep{girvan2002community}. The problem of finding such clusters~--- graph clustering~--- has been extensively studied in network science and machine learning communities, and many methods for it have been proposed over the years \citep{fortunato2010community}. Such algorithms can be used to divide graph nodes into clusters where nodes within the same cluster are strongly related, where the exact definition of this relation and thus the meaning of the obtained clusters can differ between different clustering algorithms.

In this work, we argue that various partitions of a graph into node clusters provide very useful information about the graph structure and thus it can be beneficial to integrate these partitions into GML models. Specifically, we propose \emph{cluster attention (CLATT)}~--- a graph-based attention mechanism in which nodes can attend to other nodes in the same cluster and thus exchange information with them. We view cluster attention as a middle ground between MPNNs in which only neighboring nodes can interact with each other and Graph Transformers in which all nodes can interact with each other. Cluster attention allows capturing longer-ranged dependencies than MPNNs while still providing strong graph-structure-based inductive biases which Graph Transformers lack.

We argue that the graph-structure-based inductive biases of cluster attention are complementary to those of MPNNs and thus propose augmenting MPNNs with cluster attention. Similarly, Graph Transformers can be augmented with CLATT to provide these global attention models with stronger graph-structure-based inductive biases. In experiments on a diverse set of 12 real-world graph datasets, we demonstrate that cluster attention can significantly improve the performance of both MPNNs and Graph Transformers.

The rest of this paper is structured as follows. Section~\ref{sec:background} provides the necessary background on GML with neural networks and on graph node clustering. In Section~\ref{sec:cluster-attention}, we formally define cluster attention and discuss its implementation details. In Section~\ref{sec:clustering-selection}, we discuss our approach to selecting graph node clustering algorithms to use with cluster attention. Section~\ref{sec:experiments} presents our experimental results. In Section~\ref{sec:limitations}, we discuss the limitations of cluster attention. Section~\ref{sec:conclusion} provides concluding remarks.

\section{Background and related work}
\label{sec:background}

\subsection{Graph Neural Networks and Graph Transformers}

In recent years, Graph Neural Networks (GNNs), in particular Message Passing Neural Networks (MPNNs) \citep{gilmer2017neural}, have become the most dominant approach to graph machine learning tasks. In each message passing layer of an MPNN, each node aggregates representations from its neighbors in the graph (by receiving messages from them) and updates its own representation based on the aggregated information. Many GNN architectures falling under the MPNNs framework have been proposed \citep{kipf2017semi, hamilton2017inductive, velivckovic2018graph, xu2019powerful, brody2022attentive}, mostly differing in their aggregation function. The receptive field of MPNNs, i.e., the maximum shortest path distance in the graph at which nodes can interact with each other, is equal to the number of message passing layers. This is often viewed as a limitation, since long-range dependencies can exist in some graph machine learning problems. While there have been some works attempting to increase the receptive field of MPNNs \citep{abu2019mixhop, finder2025improving}, a different and more radical approach has become more popular: replacing MPNNs with models with all-to-all Transformer-style attention \citep{vaswani2017attention} that has global receptive field. Such models became known as Graph Transformers (GTs) \citep{kreuzer2021rethinking, ying2021transformers, wu2022nodeformer, rampavsek2022recipe}. However, global attention has no information about graph topology and thus this approach foregoes graph-structure-based inductive biases of MPNNs and essentially treats the graph nodes as a set rather than a graph. To inject graph information into GTs, various positional encoding (PEs) are typically added to node features or attention weights. Such PEs are often based on Laplacian eigenvectors \citep{dwivedi2020benchmarking}, graph distances \citep{ying2021transformers}, or random walks \citep{dwivedi2022graph}. However, despite the success of the Transformer architecture in other fields such as natural language processing, computer vision, and audio processing, and despite considerable research efforts to adapt Transformers to graph machine learning, recent benchmarking works show that GTs tend to not provide advantages over MPNNs \citep{tonshoff2023did, luo2024classic, luo2025can}.

It is also important to note that there are actually two types of models commonly referred to as Graph Transformers. Besides Graph Transformers with all-to-all attention discussed above, another type consists of models that adopt the Transformer attention mechanism (and often other elements of the Transformer architecture) to GNNs but limit attention to each node's neighborhood, i.e., each node can only attend to its neighbors \citep{shi2021masked, dwivedi2021generalization, platonov2023critical}. Such models fit into the MPNNs framework and have been shown to be a very strong MPNN variant \citep{platonov2023critical, bazhenov2025graphland}. To distinguish between these two model classes, we will henceforth refer to models with all-to-all Transformer attention as Global Graph Transformers (GGTs) and to models with neighborhood Transformer attention as Local Graph Transformers (LGTs).

\subsection{Graph clustering}
\label{sec:graph-clustering}

Graph node clustering (henceforth graph clustering), also referred to as community detection, graph partitioning, and modular structure inference, is an important and long-studied problem in network science and graph machine learning with many applications in various fields. Many widely different approaches to graph clustering have been proposed, see \citet{fortunato2010community, fortunato2016community, schaeffer2007graph} for detailed overviews. Here, we briefly review some of the most well-known approaches. Many methods are based on intuitive heuristics such as iterative propagation of influence between nodes \citep{raghavan2007near} or divisive clustering by removing ``bridge-like" edges based on betweenness centrality \citep{girvan2002community} or approximate minimum cuts (and variants such as ratio cuts and normalized cuts) \citep{fiedler1973algebraic, pothen1990partitioning, shi2000normalized, meila2000learning, andersen2007using, riolo2014first}. Another approach that gained a lot of popularity is maximizing a certain measure (quality function) that shows how well-connected the nodes within a single cluster are compared to nodes from different clusters. The most well-known such measure is modularity \citep{newman2004finding, newman2006modularity}, and many algorithms to maximize it have been proposed \citep{clauset2004finding, duch2005community, reichardt2006statistical, guimera2005functional, newman2006finding}, the most popular one being the greedy Louvain algorithm \citep{blondel2008fast}. However, modularity maximization for graph clustering has some limitations \citep{fortunato2016community}, including a problem known as the resolution limit \citep{fortunato2007resolution}. An alternative to the modularity measure that does not suffer from the resolution limit is the Constant Potts Model (CPM) \citep{traag2011narrow}, which can also be maximized with the Louvain algorithm. An improvement to the Louvain algorithm that can be used to greedily maximize either modularity or CPM is the Leiden algorithm \citep{traag2019louvain}. Another popular approach to graph clustering is based on statistical inference: the graph is assumed to be generated from a certain random graph model family parametrized by community assignments of nodes, and the assignment with maximum likelihood is searched for. Methods following this approach differ in what model family they assume and how they maximize likelihood \citep{hastings2006community, newman2007mixture, zanghi2008fast, hofman2008bayesian, copic2009identifying, karrer2011stochastic, peixoto2014efficient, peixoto2014hierarchical, zhang2015identification, prokhorenkova2019community, zhang2020statistical}. The approaches discussed above consider unattributed graphs, i.e., graphs without node features. More recently, GNN-based methods for clustering attributed graphs that consider both graph structure and node features have been developed \citep{tsitsulin2023graph, liu2026bridging}.

\section{Cluster Attention}
\label{sec:cluster-attention}

\begin{figure*}
    \centering
    \begin{subfigure}[b]{0.3\textwidth}
        \centering
        \includegraphics[width=\textwidth]{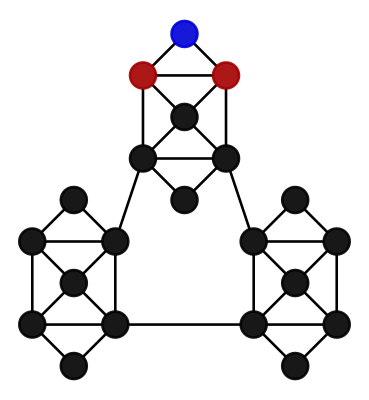}
        \caption{MPNN}
        \label{fig:receptive_field_MPNN}
    \end{subfigure}
    \hfill
    \begin{subfigure}[b]{0.3\textwidth}
        \centering
        \includegraphics[width=\textwidth]{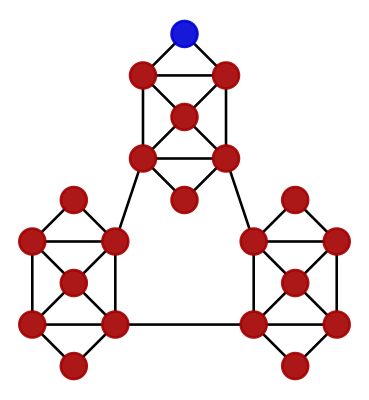}
        \caption{GGT}
        \label{fig:receptive_field_ggt}
    \end{subfigure}
    \hfill
    \begin{subfigure}[b]{0.3\textwidth}
        \centering
        \includegraphics[width=\textwidth]{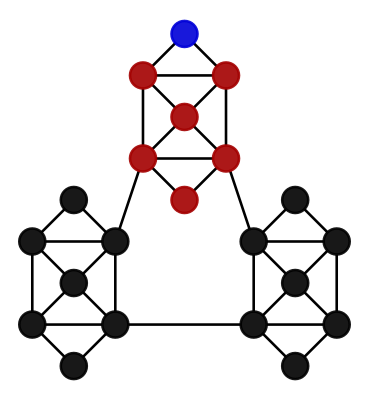}
        \caption{CLATT}
        \label{fig:receptive_field_clatt}
    \end{subfigure}
    \caption{Receptive fields of a single layer of different GML models. The blue node can interact with the red nodes. In an MPNN, the receptive field is based on the graph structure, but only neighboring nodes can interact with each other. GGT allows all nodes to interact with each other, but completely disregards the graph structure. CLATT allows nodes within the same cluster to interact, which provides longer-range interactions than in an MPNN, but still relies on the graph structure to disallow most possible pairwise interactions.}
    \label{fig:receptive_fields}
\end{figure*}

\subsection{Method}

The standard MPNN mechanism of sending messages along graph edges provides the model with strong graph-structure-based inductive biases, but prevents nodes further away in the graph than the number of message passing layers from exchanging information. We would like to design a mechanism that allows longer-range interactions while still being rooted in the graph structure and thus providing useful inductive biases to the model (in contrast to the all-to-all attention of GGTs). We propose using graph clustering to achieve this. Specifically, we apply graph clustering as a preprocessing step and then allow nodes within the same cluster to interact with each other via attention. Essentially, we use all-to-all attention on the level of individual clusters rather than on the full-graph level (as in GGTs). Such attention mechanism allows capturing longer-range dependencies than the classic message passing mechanism while still limiting which nodes can interact with each other based on the graph structure. We call this mechanism \emph{\underline{Cl}uster \underline{Att}ention (CLATT)}. We propose integrating CLATT into MPNNs alongside their message passing, since CLATT and message passing provide different kinds of graph-structure-based inductive biases that can be mutually beneficial. Similarly, CLATT can be integrated into GGTs alongside their global attention to provide GGTs with useful inductive biases. This approach is similar to how global attention of GGTs can be combined with message passing in a hybrid model \citep{rampavsek2022recipe}. In Figure~\ref{fig:receptive_fields}, we provide a simple example of how the receptive fields of a single layer of an MPNN, a GGT, and CLATT differ. Note that stacking multiple layers with message passing and CLATT allows nodes from different clusters to also interact with each other (by using several steps of message passing and/or cluster attention) thus allowing the model to capture even longer-range dependencies.

Note that modern efficient graph clustering algorithms such as the Leiden algorithm \citep{traag2019louvain} run orders of magnitude faster than it takes to train a GNN; thus, the graph clustering preprocessing step takes negligible time. However, a question remains: how to choose a graph clustering algorithm for CLATT from the many different approaches that have been proposed in the literature. We will discuss clustering algorithm selection in detail in Section~\ref{sec:clustering-selection}, but for now let us note that it is not necessary to limit CLATT to \emph{a single} graph clustering algorithm. We can select \emph{multiple} graph clusterings obtained with different algorithms, apply CLATT to each one, and then concatenate (or combine in some other way, e.g., via summation) the resulting representations for each node. Different clustering algorithms rely on different assumptions and have different inductive biases, thus producing significantly different clusterings that capture different types of information about the graph structure, and using multiple such clusterings lets a model access all of these different types of information.

Now let us describe CLATT more formally. Let $G = (V, E)$ be a graph with the nodeset $V$ and the edgeset $E$. Let $\mathcal{C}$ be an ordered set of clusterings that contains one or more clusterings that we want to use with CLATT. For each clustering $C \in \mathcal{C}$ and each node $i \in V$ we will denote by $C(i)$ the cluster (the set of nodes) to which the node $i$ belongs in the clustering $C$. Let $\mathbf{x}_i \in \mathbb{R}^d$ be the representation of node $i \in V$ before the CLATT operation, i.e., the input to CLATT, and let $\mathbf{y}_i \in \mathbb{R}^{d |\mathcal{C}|}$ be the representation of node $i \in V$ after the CLATT operation, i.e., the output of CLATT. The complete CLATT output representation $\mathbf{y}_i \in \mathbb{R}^{d |\mathcal{C}|}$ for node $i$ is the concatenation of CLATT output representations $\mathbf{y}_i^C \in \mathbb{R}^d$ for each clustering $C \in \mathcal{C}$:
\begin{equation}
\mathbf{y}_i = \mathrm{concatenate} \left( \mathbf{y}_i^C \;\, \mathrm{for} \;\, C \;\, \mathrm{in} \;\, \mathcal{C} \right).
\end{equation}
The CLATT output representation $\mathbf{y}_i^C$ for a single clustering $C \in \mathcal{C}$ for node $i$ is computed via attention between nodes belonging to the cluster $C(i)$ in the following way (for ease of notation we assume a single attention head, but we use multihead attention in practice):
\begin{equation}
\mathbf{y}_i^C = \sum_{j \in C(i)} p_{ij}^C \mathbf{v}_j^C,
\end{equation}
where $p_{ij}^C \in [0, 1]$ is the weight (probability) of attention from node $i$ to node $j$ obtained with softmax as
\begin{equation}
p_{ij}^C = \frac{ \exp \left( \alpha_{ij}^C \right) }{\sum_{j \in C(i)} \exp \left( \alpha_{ij}^C \right) },
\end{equation}
$\alpha_{ij}^C$ is the pre-softmax logit of attention from node $i$ to node $j$ obtained as a scaled dot product between the corresponding query and key vectors:
\begin{equation}
\alpha_{ij}^C = \frac{ \langle \mathbf{q}_i^C, \mathbf{k}_j^C \rangle }{\sqrt{d}},
\end{equation}
and $\mathbf{q}_i^C, \mathbf{k}_i^C, \mathbf{v}_i^C \in \mathbb{R}^d$  are the query, key, and value vectors respectively for node $i \in V$ obtained as learnable linear transformations of the node's representation:
\begin{equation}
\mathbf{q}_i^C = \mathbf{W}_q^C \mathbf{x}_i + \mathbf{b}_q^C, \;\;\;\;\;\;\;\;\;\; \mathbf{k}_i^C = \mathbf{W}_k^C \mathbf{x}_i + \mathbf{b}_k^C, \;\;\;\;\;\;\;\;\;\; 
\mathbf{v}_i^C = \mathbf{W}_v^C \mathbf{x}_i + \mathbf{b}_v^C,
\end{equation}
where $\mathbf{W}_q^C, \mathbf{W}_k^C, \mathbf{W}_v^C \in \mathbb{R}^{d \times d}$ and $\mathbf{b}_q^C, \mathbf{b}_k^C, \mathbf{b}_v^C \in \mathbb{R}^d$ are learnable parameters. The complete CLATT output representation $\mathbf{y}_i \in \mathbb{R}^{d |\mathcal{C}|}$ for node $i$ is then concatenated with the message passing output (in a CLATT-augmented MPNN model) or the global attention output (in a CLATT-augmented GGT model) for node $i$ (which is typically of dimension $d$) and passed through a learnable linear layer that reduces its dimension back to $d$.

\subsection{Implementation details}

CLATT can be naively implemented by adding edges connecting all pairs of nodes within the same cluster to the graph. However, this will likely create a very dense graph for which graph attention operation meant for sparser graphs will be inefficient. Instead, we transform the tensor of node representations of shape \hbox{\texttt{num\_nodes $\times$ hidden\_dim}} to a padded tensor of shape \hbox{\texttt{num\_clusters $\times$ max\_cluster\_size $\times$ hidden\_dim}} and apply standard dense attention along the \texttt{max\_cluster\_size} dimension with a mask that blocks attention to padding. Modern efficient attention implementations\footnote{\url{https://docs.pytorch.org/docs/stable/generated/torch.nn.functional.scaled_dot_product_attention.html}} combined with JIT-compilation\footnote{\url{https://docs.pytorch.org/tutorials/intermediate/torch_compile_tutorial.html}} make this operation very efficient. Further, some modern frameworks provide so-called ``ragged / jagged'' tensors\footnote{\url{https://docs.pytorch.org/docs/stable/nested.html}}\footnote{\url{https://docs.pytorch.org/tutorials/unstable/nestedtensor.html}} which can be used to stack sequences of different length into a single tensor-like structure, which can allow avoiding padding to make CLATT implementation even more efficient.

\section{Clustering algorithms selection}
\label{sec:clustering-selection}

Considering the wide and diverse range of graph clustering algorithms available (see Section~\ref{sec:graph-clustering}), it is important to choose appropriate clustering algorithms to use with CLATT. Different clustering algorithms rely on different principles and make different assumptions about the data, and thus can produce significantly different clusterings. Since real-world graphs come from many different applications and have very different structures, the best clustering algorithm for use with CLATT can also be different for different datasets and tasks. First, we note that we do not need to limit CLATT to using \emph{a single} clustering: we can select \emph{multiple} clustering algorithms and apply CLATT to clusterings obtained from each of them. However, we still need to choose a relatively small set of clustering algorithms to use from the wide range of available ones. In this selection, we use the following two criteria: first, we only consider efficient algorithms that can be run relatively fast on large-scale real-world graphs, making graph clustering a cheap preprocessing step compared to GNN training; second, we aim to select a set of clustering algorithms that produce significantly different results compared to each other, thus providing CLATT with different information about the graph structure when used together. Based on these criteria, we select 4 clustering algorithms. In practice, different clustering algorithms can have different usefulness for different datasets and tasks, and there is no need to always use all 4 of them. Thus, we treat the choice of clustering algorithms for each dataset as a hyperparameter. First, we train a model 4 times with each clustering individually; then, we take those of the considered clusterings that improved results on the validation set and use them together for the main training run. To speed up experiments, we only run this procedure with one of the considered models (specifically, LGT) and use the same selection of clusterings for all other models (we assume that the selection of clusterings will transfer well between different models, but it is possible that even better results can be achieved if this selection is run for each model individually).

Now, let us describe the 4 clustering algorithms that we use.

\begin{itemize}[leftmargin=10pt]

\item First, we consider the widely used Leiden algorithm \citep{traag2019louvain}~--- an improvement of the classic Louvain algorithm \citep{blondel2008fast}. Leiden algorithm greedily optimizes a measure of how well-connected nodes within a single cluster are compared to nodes from different clusters. As this measure, we use the Constant Potts Model (CPM) \citep{traag2011narrow}, which does not suffer from the resolution limit problem \citep{fortunato2007resolution}, in contrast to the more widely used modularity measure \citep{newman2004finding, newman2006modularity}. We use the official algorithm implementation.\footnote{\url{leidenalg.readthedocs.io}}

\item Next, we consider a statistical clustering algorithm. Specifically, we use an algorithm that fits a Bayesian planted partition model by \citet{zhang2020statistical}. We use the official algorithm implementation from the \texttt{graph-tool} library \citep{peixoto2017graph}.

\item Note that the two algorithms discussed above assume \emph{assortative} community structure, i.e., nodes within the same cluster are more likely to be connected than nodes from different clusters. While such community structure is natural, sometimes it can also be useful to find \emph{disassortative} clusters, i.e., clusters of nodes that are not necessarily connected with each other but share the same structural role in the graph. For example, when a lot of leaf nodes are connected to one node in the graph, they can be expected to share some properties, and it can be useful to put them in a cluster of their own even though they are not connected with each other. Thus, we also consider a statistical clustering algorithm that does not make the assumption of cluster assortativity and can detect disassortative clusters. Specifically, we use an algorithm by \citet{peixoto2014hierarchical}. This algorithm produces a hierarchical clustering; however, we only consider the level of the hierarchy with the smallest nontrivial number of clusters as we find that other levels typically find clusters that are too small. We use the official algorithm implementation from the \texttt{graph-tool} library \citep{peixoto2017graph}.

\item Above, we have described 3 very different graph clustering algorithms. However, it can sometimes be useful to also consider the \emph{feature similarity} of nodes, i.e., ignore the graph structure and apply a classic clustering algorithm to node features. This approach will allow the model to exchange information between nodes that have similar features no matter how far they are in the graph, which can be useful for some applications. For this purpose, we use the classic k-means algorithm \citep{forgy1965cluster, lloyd1982least}. We use the implementation from the \texttt{scikit-learn} library \citep{pedregosa2011scikit}. However, since the k-means algorithm relies on euclidean distances between feature vectors, it is not particularly suitable for datasets with heterogeneous features of different scales and distributions, which are common in practical GML applications \citep{bazhenov2025graphland}. Thus, instead of directly clustering node features, we train an auxiliary ResMLP model (a simple graph-agnostic model that is very fast to train) on the dataset, obtain hidden representations of graph nodes from this model, and cluster these representations. We believe it is interesting to compare this feature-based clustering to graph-structure-based clusterings. In our experiments, we find that this feature-based clustering rarely helps (see Appendix~\ref{app:clustering-algorithms}), which confirms our intuition that graph-structure-based inductive biases are more useful for GML.

\end{itemize}

\begin{figure*}[t]
    \centering
    \includegraphics[width=\textwidth]{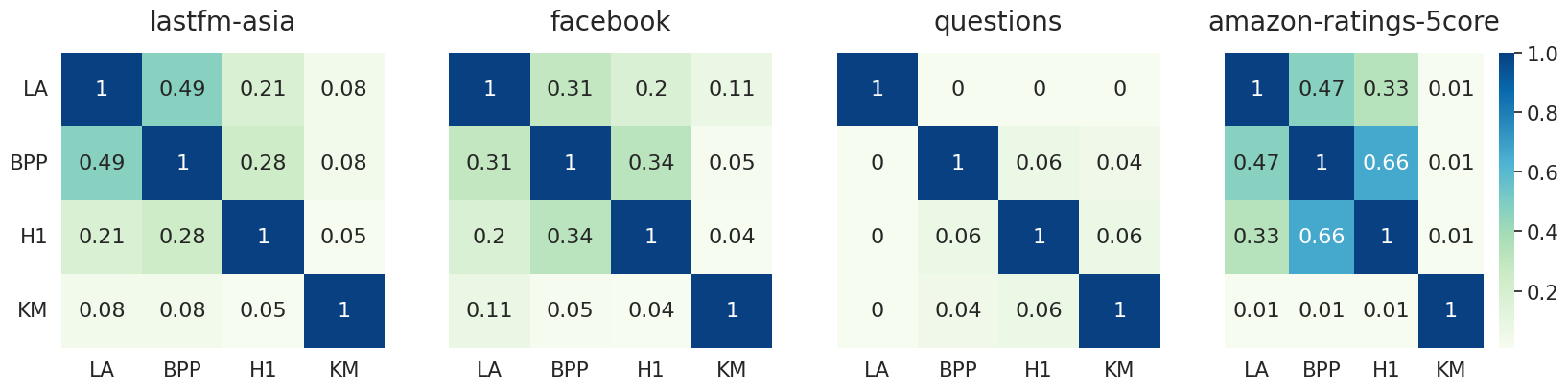}
    \caption{Similarities of clusterings produced by different node clustering algorithms on 4 different datasets: \texttt{lastfm-asia}, \texttt{facebook}, \texttt{questions}, \texttt{amazon-ratings-5core}. The Correlation Coefficient similarity measure is reported. LA is Leiden algorithm, BPP is Bayesian planted partition model, H1 is hierarchical statistical clustering (with the hierarchy level with the smallest non-trivial number of clusters being used), KM is k-means clustering of ResMLP node representations.}
    \label{fig:clusterings-similarities}
\end{figure*}

The 4 considered algorithms are very efficient and also typically produce significantly different clusterings, thus satisfying our desiderata. To verify that these algorithms produce significantly different results, we can compare the clusterings obtained from these algorithms using some \emph{clustering similarity measure}. While many such measures have been proposed in the literature \citep{gosgens2021systematic}, we use the measure known as \emph{Correlation Coefficient (CC)} since it has been shown to satisfy many desirable properties \citep{gosgens2021systematic}. We show the similarity matrices of the clusterings obtained by the 4 considered algorithms on 4 datasets used in our experiments (see Appendix~\ref{app:datasets} for dataset descriptions) in Figure~\ref{fig:clusterings-similarities}. Indeed, the similarity values of clusterings are mostly relatively low, with the graph-agnostic k-means clustering typically being particularly dissimilar to the other (graph-based) clusterings, as expected.

\begin{table*}[t]
\caption{Characteristics of the considered graph datasets.}
\label{tab:datasets-characteristics}
\setlength\tabcolsep{5pt}
\centering
\input{inputs/tables/dataset_characteristics}
\end{table*}

\section{Experiments}
\label{sec:experiments}

\subsection{Datasets}
\label{sec:datasets}

In our experiments, we aim to show that CLATT can improve results on diverse datasets from different domains. In particular, we aim to go beyond the classic citation network datasets, extensive reliance on which has been criticized recently \citep{bechler2025position, bazhenov2025graphland, platonov2026fair}, and use graph datasets from more realistic and practically relevant applications. Thus, we select 12 datasets from different fields and with different tasks, showcasing a wide range of graph sizes and structural characteristics. First, we use 7 datasets from the recently introduced GraphLand benchmark \citep{bazhenov2025graphland} that focuses on graph datasets from realistic industrial applications and with rich node features. Specifically, we use road networks \texttt{city-roads-M} and \texttt{city-roads-L}, a network of internet-connected devices \texttt{avazu-ctr}, a network of crowdsourcing platform workers \texttt{tolokers-2}, a network of review service users \texttt{city-reviews}, a co-purchasing network \texttt{hm-categories}, and a social network \texttt{pokec-regions}. For datasets from GraphLand, we use the official \texttt{RL} (random low) 10\%/10\%/80\% train/val/test data splits. Further, we use two classic social network datasets \texttt{lastfm-asia} \citep{rozemberczki2020feather} and \texttt{facebook} \citep{rozemberczki2019gemsec}. For these two datasets, we use random 10\%/10\%/80\% train/val/test data splits. Then, we use two datasets from \citet{platonov2023critical}: co-purchasing network \texttt{amazon-ratings} and question-answering network \texttt{questions}. For these two datasets, we use random 50\%/25\%/25\% train/val/test data splits. Finally, we notice that the \texttt{amazon-ratings} dataset has a peculiar graph structure due to being downsampled by taking the 5-core \citep{malliaros2020core} of the full graph. For this reason, we further refer to this dataset as \texttt{amazon-ratings-5core}, and we also prepare a new dataset which is the full version of the same co-purchasing network, which we call \texttt{amazon-ratings-full}. Besides being more than an order of magnitude larger, this dataset exhibits very different structural characteristics compared to \texttt{amazon-ratings-5core}. For this dataset, we use a random 10\%/10\%/80\% train/val/test data split. A detailed description of this new dataset and how it differs from \texttt{amazon-ratings-5core} is provided in Appendix~\ref{app:amazon-ratings-full}.

To showcase the diversity of graph structures in the considered datasets, we provide some of their structural characteristics in Table~\ref{tab:datasets-characteristics}. More details on these datasets and their characteristics are provided in Appendix~\ref{app:datasets}. For example, note that we use a range of both homophilous datasets (\texttt{city-roads-M}, \texttt{city-roads-L}, \texttt{city-reviews}, \texttt{pokec-regions}, \texttt{lastfm-asia}, \texttt{facebook}) and non-homophilous datasets (\texttt{avazu-ctr}, \texttt{tolokers-2}, \texttt{hm-categories}, \texttt{questions}, \texttt{amazon-ratings-5core}, \texttt{amazon-ratings-full}) to demonstrate that CLATT works well for both of these types of datasets.

The largest of the considered datasets~--- \texttt{pokec-regions} --- has more than $1.6$ million nodes. GGTs are not efficient enough to handle datasets of such size in reasonable time (a single training run of GGT on this dataset takes approximately $65$ hours using an NVIDIA Tesla A100 80GB GPU), while our LGT-CLATT model (which is a CLATT-augmented version of the most computationally expensive of the considered MPNNs) can be trained on it in approximately half an hour (using the same GPU), which demonstrates the efficiency of our CLATT implementation.

\subsection{Models}

As discussed above, we propose integrating CLATT into MPNNs to combine the benefits of local message passing and more long-range cluster attention. We combine CLATT with three different MPNN models and show that CLATT improves results for all of them. Specifically, we use two classic GNNs --- GCN \citep{kipf2017semi} and GraphSAGE \citep{hamilton2017inductive} --- and also Local Graph Transformer (LGT), which has been shown to often provide stronger results in recent benchmarking works \citep{platonov2023critical, bazhenov2025graphland}. For all models, we use the modifications from \citet{platonov2023critical} which augment models with skip-connections \citep{he2016deep}, layer normalization \citep{ba2016layer}, and MLP blocks between message passing layers. These modifications have been shown to significantly improve the results of MPNNs \citep{luo2024classic, luo2025can}.

Further, we also combine CLATT with Global Graph Transformers (GGTs) and show that it improves their performance as well. We experiment with two versions of GGT that differ by the positional encodings used: we use either classic DeepWalk node embeddings \citep{perozzi2014deepwalk} or Laplacian eigenvectors \citep{dwivedi2020benchmarking, belkin2001laplacian}. We denote these models \hbox{GGT-DW and GGT-Lap}, respectively. Note that while many alternative node embedding methods have been proposed since DeepWalk, it has been shown that in practice DeepWalk still provides some of the best results \citep{gurukar2022benchmarking}.

For all models, we perform extensive hyperparameter search. The details of it, as well as other information about our experimental setting, are provided in Appendix~\ref{app:experimental-setup}.

\begin{table*}[h!]
\caption{Experimental results. $R^2$ is reported for regression datasets, Average Precision is reported for binary classification datasets, Accuracy is reported for multiclass classification datasets. TL means time limit exceeded (24 hours for a single model run). For each pair consisting of a base model and its version augmented with CLATT, we highlight the best of the two results with bold and additionally underline it if the difference between the two results is statistically significant.}
\label{tab:results}
\begin{subtable}[t]{\textwidth}
\caption{Experimental results for GraphLand datasets under the \texttt{RL} data split.}
\centering
\input{inputs/tables/results_graphland}
\label{tab:results-graphland}
\end{subtable}
\begin{subtable}[t]{\textwidth}
\vspace{8pt}
\caption{Experimental results for other datasets.}
\centering
\input{inputs/tables/results_other}
\label{tab:results-other}
\end{subtable}
\end{table*}

\subsection{Experimental results}

We report the results of our experiments in Table~\ref{tab:results}. For each of the considered models, we compare the base version of the model with its version augmented with CLATT (which is denoted by -CLATT suffix). We can see that CLATT improves the performance of the base models in all cases with the only exception of the GraphSAGE model on the \texttt{amazon-ratings-5core} dataset, and these improvements are often quite substantial. For example, on the \texttt{pokec-regions} dataset, CLATT improves the performance of GCN and GraphSAGE by more than 13 and 10 percentage points, respectively. The improvements from CLATT are not limited to MPNNs: for GGTs, CLATT always improves performance by providing these models with graph-structure-based inductive biases. For example, on the \texttt{hm-categories} dataset, CLATT improves the performance of GGT-DW and GGT-Lap by more than 13 and 15 percentage points, respectively.

We note that CLATT brings improvements on datasets with very different graph structural characteristics. As discussed above, the considered datasets cover both homophilous and non-homophilous graphs. Further, CLATT improves model performance on both very dense graphs (\texttt{avazu-ctr}, \texttt{hm-categories}) and very sparse graphs (\texttt{city-roads-M}, \texttt{city-roads-L}, \texttt{questions}, \texttt{amazon-ratings-full}). Note that \texttt{city-roads-M} and \texttt{city-roads-L} datasets, representing transportation networks, do not exhibit particularly well-defined clusters (as can be expected given their zero clustering coefficients and very large average distances), in contrast to, for example, social networks, yet CLATT brings strong benefits on them as well. More details on the diversity of graph structural characteristics covered by the considered datasets are provided in Appendix~\ref{app:datasets}.

In Appendix~\ref{app:attention}, we provide an analysis of attention patterns learned by local, global, and cluster attention. It demonstrates that, on the one hand, cluster attention often works at longer ranges than the receptive field of local attention models permits, which allows it to capture useful long-range dependencies not available to local attention, but, on the other hand, cluster attention typically works at shorter ranges than global attention, which struggles to leverage useful information encoded in the graph topology due to its lack of appropriate graph-structure-based inductive biases.

Specific clustering algorithms selected during hyperparameter search for each of the considered datasets are provided in Appendix~\ref{app:clustering-algorithms}. We can see that using CLATT with more than one clustering algorithm simultaneously is beneficial for 50\% of the datasets. The Leiden algorithm is the most frequently selected among the considered algorithms (it turns out to be helpful for 75\% of the datasets), but the two considered statistical clustering algorithms are also regularly selected. In contrast, the k-means algorithm applied to ResMLP node representations turns out to be helpful for only one of the considered datasets (\texttt{questions}). This shows that using graph-structure-based clusterings with CLATT is much more beneficial than using feature-similarity-based clusterings, which agrees with our intuition that CLATT is most useful for providing additional graph-structure-based inductive biases to GML models.

Finally, we observe that MPNNs almost always outperform GGTs on the considered datasets (while also being much more efficient). This is in agreement with recent works \citep{tonshoff2023did, luo2024classic, luo2025can} showing that MPNNs perform better than GGTs (note that the majority of our datasets are different and more realistic and practically relevant than those used in these works). However, CLATT tends to help GGTs more than MPNNs (due to GGTs not having strong enough graph-structure-based inductive biases without it), allowing GGTs to partially close the gap.

\section{Limitations}
\label{sec:limitations}

Graphs can be used to represent data from very different domains, and thus can exhibit a wide range of structural characteristics and relationships between graph structure, node features, and node labels. Thus, we do not expect that a single method will perform well on all realistic graphs. Yet, we show that CLATT can improve the performance of both MPNNs and GGTs (two currently most popular families of neural architectures for graph machine learning) on a wide range of graph datasets from real-world applications of graph machine learning. However, counterexamples of graphs on which CLATT does not improve performance are of course possible. We expect that the effectiveness of CLATT is correlated with the degree of the presence of meaningful cluster structure in the graph. While many graphs exhibit easily detectable cluster structure, some graphs, for example, highly regular grid-like graphs, are unlikely to contain meaningful clusters and thus probably will not benefit from CLATT. However, many real-world networks do contain useful cluster structure that CLATT can leverage to improve model performance, which we demonstrate in our experiments.

\section{Conclusion}
\label{sec:conclusion}

In this work, we introduce a new technique for graph machine learning --- cluster attention (CLATT), which can be viewed as a middle ground between local message passing and global attention that incorporates strong graph-structure-based inductive biases while allowing for long-range interactions. We show that CLATT significantly improves the results of both MPNNs and GGTs on a wide range of diverse real-world graph datasets.

\bibliography{neurips_2026}
\bibliographystyle{refstyle}

\newpage

\appendix

\section{Attention distances analysis}
\label{app:attention}

In this section, we investigate how far nodes typically look with different types of attention. Specifically, we analyze the average distances between each node and the nodes it attends to weighted with attention probabilities for the three considered graph attention mechanisms: local (1-hop neighborhood), global (full-graph), and cluster attention. We consider three base models~--- LGT (local attention), GGT-DW (global attention), GGT-Lap (global attention), and their cluster attention augmented variants --- LGT-CLATT, GGT-DW-CLATT, GGT-Lap-CLATT, and analyze their attention patterns on two datasets --- \texttt{lastfm-asia} and \texttt{tolokers-2}. We limit ourselves to these two datasets because they are the smallest among the ones considered in our work, and our analysis requires the full attention matrices for global attention, which require memory quadratic in the number of nodes (in our main experiments, we use efficient FlashAttention-based \citep{dao2022flashattention} attention algorithms which do not materialize the full attention matrices and thus allow us to scale even to very large graphs; however, for the analysis in this section, we need the full attention matrices).

\begin{figure*}
    \centering
    \begin{subfigure}[b]{0.49\textwidth}
        \centering
        \caption*{\large lastfm-asia}
        \vspace{12pt}
        \renewcommand{\thesubfigure}{a1}
        \includegraphics[width=0.8\textwidth]{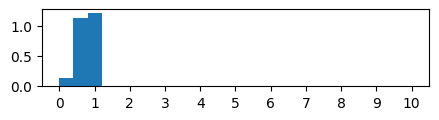}
        \vspace{-6pt}
        \caption{LGT (local attention)}
        \vspace{12pt}
    \end{subfigure}
    \hfill
    \begin{subfigure}[b]{0.49\textwidth}
        \centering
        \caption*{\large tolokers-2}
        \vspace{12pt}
        \renewcommand{\thesubfigure}{b1}
        \includegraphics[width=0.8\textwidth]{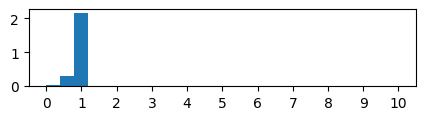}
        \vspace{-6pt}
        \caption{LGT (local attention)}
        \vspace{12pt}
    \end{subfigure}
    \hfill
    \begin{subfigure}[b]{0.49\textwidth}
        \centering
        \renewcommand{\thesubfigure}{a2}
        \includegraphics[width=0.8\textwidth]{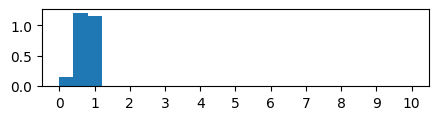}
        \vspace{-6pt}
        \caption{LGT-CLATT (local attention)}
        \vspace{12pt}
    \end{subfigure}
    \hfill
    \begin{subfigure}[b]{0.49\textwidth}
        \centering
        \renewcommand{\thesubfigure}{b2}
        \includegraphics[width=0.8\textwidth]{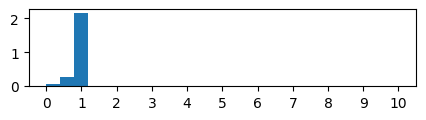}
        \vspace{-6pt}
        \caption{LGT-CLATT (local attention)}
        \vspace{12pt}
    \end{subfigure}
    \hfill
    \begin{subfigure}[b]{0.49\textwidth}
        \centering
        \renewcommand{\thesubfigure}{a3}
        \includegraphics[width=0.8\textwidth]{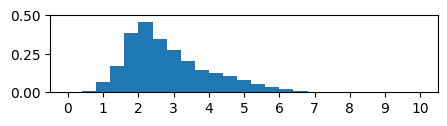}
        \vspace{-6pt}
        \caption{LGT-CLATT (cluster attention)}
        \vspace{12pt}
    \end{subfigure}
    \hfill
    \begin{subfigure}[b]{0.49\textwidth}
        \centering
        \renewcommand{\thesubfigure}{b3}
        \includegraphics[width=0.8\textwidth]{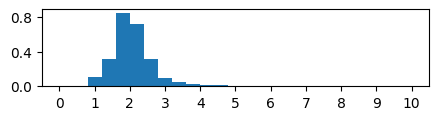}
        \vspace{-6pt}
        \caption{LGT-CLATT (cluster attention)}
        \vspace{12pt}
    \end{subfigure}
    \hfill
    \begin{subfigure}[b]{0.49\textwidth}
        \centering
        \renewcommand{\thesubfigure}{a4}
        \includegraphics[width=0.8\textwidth]{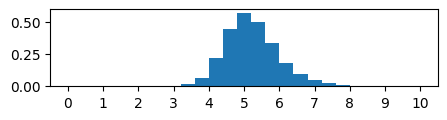}
        \vspace{-6pt}
        \caption{GGT-DW (global attention)}
        \vspace{12pt}
    \end{subfigure}
    \hfill
    \begin{subfigure}[b]{0.49\textwidth}
        \centering
        \renewcommand{\thesubfigure}{b4}
        \includegraphics[width=0.8\textwidth]{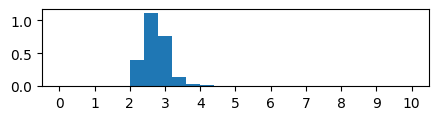}
        \vspace{-6pt}
        \caption{GGT-DW (global attention)}
        \vspace{12pt}
    \end{subfigure}
    \hfill
    \begin{subfigure}[b]{0.49\textwidth}
        \centering
        \renewcommand{\thesubfigure}{a5}
        \includegraphics[width=0.8\textwidth]{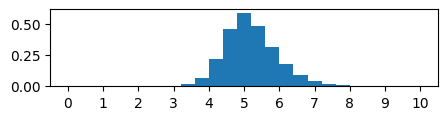}
        \vspace{-6pt}
        \caption{GGT-DW-CLATT (global attention)}
        \vspace{12pt}
    \end{subfigure}
    \hfill
    \begin{subfigure}[b]{0.49\textwidth}
        \centering
        \renewcommand{\thesubfigure}{b5}
        \includegraphics[width=0.8\textwidth]{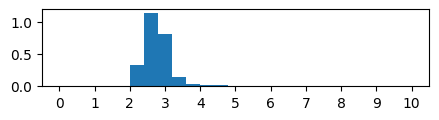}
        \vspace{-6pt}
        \caption{GGT-DW-CLATT (global attention)}
        \vspace{12pt}
    \end{subfigure}
    \hfill
    \begin{subfigure}[b]{0.49\textwidth}
        \centering
        \renewcommand{\thesubfigure}{a6}
        \includegraphics[width=0.8\textwidth]{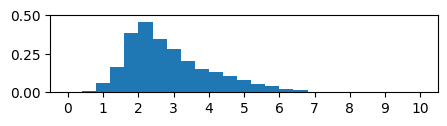}
        \vspace{-6pt}
        \caption{GGT-DW-CLATT (cluster attention)}
        \vspace{12pt}
    \end{subfigure}
    \hfill
    \begin{subfigure}[b]{0.49\textwidth}
        \centering
        \renewcommand{\thesubfigure}{b6}
        \includegraphics[width=0.8\textwidth]{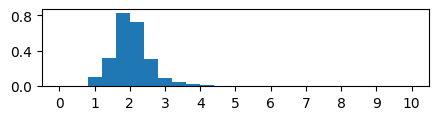}
        \vspace{-6pt}
        \caption{GGT-DW-CLATT (cluster attention)}
        \vspace{12pt}
    \end{subfigure}
    \hfill
    \begin{subfigure}[b]{0.49\textwidth}
        \centering
        \renewcommand{\thesubfigure}{a7}
        \includegraphics[width=0.8\textwidth]{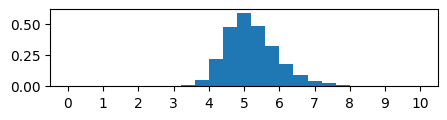}
        \vspace{-6pt}
        \caption{GGT-Lap (global attention)}
        \vspace{12pt}
    \end{subfigure}
    \hfill
    \begin{subfigure}[b]{0.49\textwidth}
        \centering
        \renewcommand{\thesubfigure}{b7}
        \includegraphics[width=0.8\textwidth]{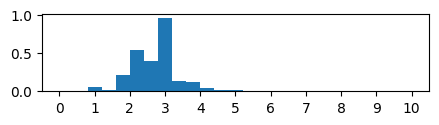}
        \vspace{-6pt}
        \caption{GGT-Lap (global attention)}
        \vspace{12pt}
    \end{subfigure}
    \hfill
    \begin{subfigure}[b]{0.49\textwidth}
        \centering
        \renewcommand{\thesubfigure}{a8}
        \includegraphics[width=0.8\textwidth]{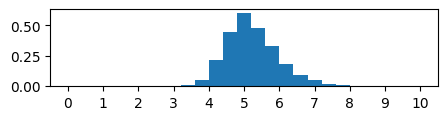}
        \vspace{-6pt}
        \caption{GGT-Lap-CLATT (global attention)}
        \vspace{12pt}
    \end{subfigure}
    \hfill
    \begin{subfigure}[b]{0.49\textwidth}
        \centering
        \renewcommand{\thesubfigure}{b8}
        \includegraphics[width=0.8\textwidth]{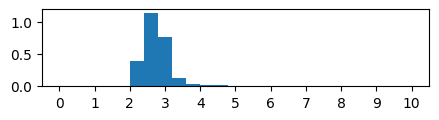}
        \vspace{-6pt}
        \caption{GGT-Lap-CLATT (global attention)}
        \vspace{12pt}
    \end{subfigure}
    \hfill
    \begin{subfigure}[b]{0.49\textwidth}
        \centering
        \renewcommand{\thesubfigure}{a9}
        \includegraphics[width=0.8\textwidth]{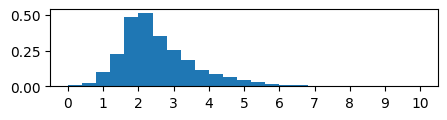}
        \vspace{-6pt}
        \caption{GGT-Lap-CLATT (cluster attention)}
        \vspace{12pt}
    \end{subfigure}
    \hfill
    \begin{subfigure}[b]{0.49\textwidth}
        \centering
        \renewcommand{\thesubfigure}{b9}
        \includegraphics[width=0.8\textwidth]{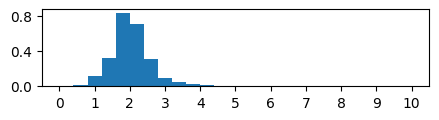}
        \vspace{-6pt}
        \caption{GGT-Lap-CLATT (cluster attention)}
        \vspace{12pt}
    \end{subfigure}
    \caption{Distributions of average attention distances for different types of attention. Note that the unweighted average pairwise distance is $5.23$ for \texttt{lastfm-asia} and $2.79$ for \texttt{tolokers-2}.}
    \label{fig:attention-distances-distributions}
\end{figure*}

For each dataset, each model, each attention type in the model (local, global, cluster attention), and each attention head, we compute the average distance to the nodes to which each node attends weighted by the corresponding attention probabilities, thus obtaining $nk$ distances per dataset, model, and attention type, where $n$ is the number of nodes in the graph and $k$ is the number of attention heads corresponding to this attention type in all layers of the model combined (for cluster attention on the \texttt{lastfm-asia} dataset, there are actually $3nk$ distances since 3 different clusterings are used). In Figure~\ref{fig:attention-distances-distributions}, we provide histograms of distributions of these average attention distances, and in Table~\ref{tab:attention-distances-distributions}, we report the $0.05$, $0.25$, $0.50$, $0.75$, $0.95$ quantiles of these distributions. Note that for local attention, the average attention distance is upper bounded by $1$, since nodes can only attend to themselves (at a distance of $0$) and to their direct neighbors in the graph (at a distance of $1$).

First, let us note that the attention distance distributions for each of the 3 attention types are significantly different between the \texttt{lastfm-asia} and \texttt{tolokers-2} datasets. For global and cluster attention, this can be explained by the very different graph structure of these two datasets: in \texttt{lastfm-asia}, the unweighted average pairwise distance between nodes is a lot larger than in \texttt{tolokers-2}, and thus attention also typically works at greater distances in \texttt{lastfm-asia} than in \texttt{tolokers-2}. However, we note that local attention also has different distance distributions for these two datasets: in \texttt{lastfm-asia} nodes attend to themselves (rather than their neighbors) in local attention significantly more than in \texttt{tolokers-2}.

Now, let us focus on comparing cluster attention to other types of attention. We can see that in cluster attention, the vast majority of nodes attend on average at significantly larger distances than in local attention (in which the maximum distance is limited to $1$), and often even at larger distances than the receptive field of a multilayer local-attention MPNN allows (which for this experiment is a distance of 3 since 3-layer MPNNs are used). Specifically, in \texttt{lastfm-asia}, more than $25$\% of nodes attend in cluster attention at distances larger than $3.5$ on average, while in \texttt{tolokers-2}, more than $5$\% of nodes attend in cluster attention at distances larger than $3.1$ on average, which are distances unattainable for local attention even with $3$ layers. However, in cluster attention nodes also on average attend at significantly smaller distances than in global attention. For example, in \texttt{lastfm-asia}, the $0.75$ quantile of cluster attention distance distributions is smaller than even the $0.05$ quantile of global attention distance distributions for all model variants, which means that at least $75\%$ of nodes in cluster attention attend on average at a distance smaller than that $95\%$ of nodes in global attention attend on average at. The situation is similar for \texttt{tolokers-2}, although a bit less drastic, since distances are generally smaller in this dataset. Overall, it can be seen that global attention struggles to focus on nearby nodes in the graph, with the vast majority of nodes in global attention attending on average at a distance of more than $4$ in \texttt{lastfm-asia} and more than $2$ in \texttt{tolokers-2}, while cluster attention gives much larger focus to nearby nodes. This shows that, despite the use of positional encodings, global attention struggles to leverage useful information about node relations encoded in the graph structure and does not focus enough on graph neighbors, which explains why integrating CLATT into GGTs improves their performance significantly.

Our results from Section~\ref{sec:experiments} show that adding cluster attention consistently improves the performance of both local-attention and global-attention models, and our observations from this section provide an explanation for why this happens. It can be useful for a model to pass information beyond the receptive field of local-attention models (and other MPNNs), which can be achieved with cluster attention. While global-attention models also provide an opportunity for using long-range interactions, due to their lack of graph-structure-based inductive biases, they struggle to focus on nearby nodes and often use attention to attend further in the graph than is actually needed, which is why they also benefit from cluster attention which has strong graph-structure-based inductive biases and allows these models to more effectively focus on nodes close in the graph.

\begin{table*}
\caption{Quantiles of distributions of average attention distances for different types of attention. Note that the unweighted average pairwise distance is $5.23$ for \texttt{lastfm-asia} and $2.79$ for \texttt{tolokers-2}.}
\label{tab:attention-distances-distributions}
\centering
\input{inputs/tables/attention}
\end{table*}

\section{Clustering algorithms selected}
\label{app:clustering-algorithms}

\begin{table*}
\caption{Clustering algorithms selected for each of the considered datasets. LA is Leiden algorithm, BPP is Bayesian planted partition model, H1 is hierarchical statistical clustering (with the hierarchy level with the smallest non-trivial number of clusters being used), KM is k-means clustering of ResMLP node representations.}
\label{tab:clustering-algorithms}
\centering
\input{inputs/tables/clustering_algorithms}
\end{table*}

As described in Section~\ref{sec:clustering-selection}, for each dataset, during hyperparameter search we select a subset of the 4 considered clustering algorithms based on the validation set performance. Table~\ref{tab:clustering-algorithms} specifies which clustering algorithms were selected for each of the considered datasets. All the considered algorithms except for k-means use the graph structure to cluster graph nodes, while k-means is applied to node representations obtained from a graph-agnostic ResMLP model and thus uses a variant of feature similarity (i.e., similarity of features transformed by a ResMLP). It can be seen that for 6 datasets (i.e., $50\%$ of the considered datasets), only one clustering algorithm is selected, for another 5 datasets, two clustering algorithms are selected, and for 1 dataset (\texttt{lastfm-asia}), three clustering algorithms are selected. The Leiden algorithm is the most frequently selected clustering algorithm: it is selected for $9$ datasets, i.e., $75\%$ of the considered datasets. The two considered statistical clustering algorithms are also regularly selected: the Bayesian planted partition model is selected for 5 datasets, while the hierarchical clustering algorithm that does not assume assortative clustering structure is selected for 4 datasets. In contrast, the k-means algorithm is the least frequently selected algorithm~--- it is only selected once (for the \texttt{questions} dataset). This shows that algorithms that use the graph structure to cluster graph nodes are typically more useful for CLATT than algorithms that use node feature similarity, which agrees with our intuition that CLATT is most beneficial in graph machine learning when used to provide additional graph-structure-based inductive biases.

\section{Datasets}
\label{app:datasets}

\subsection{Data splits}

In our experiments, we aim to use a range of datasets from different real-world applications of graph machine learning. Brief descriptions of these datasets are provided in Section~\ref{sec:datasets}. For datasets from the GraphLand benchmark \citep{bazhenov2025graphland} (\texttt{city-roads-M}, \texttt{city-roads-L}, \texttt{avazu-ctr}, \texttt{tolokers-2}, \texttt{city-reviews}, \texttt{hm-categories}, \texttt{pokec-regions}) we use the official \texttt{RL} (random low) data splits, which are random stratified 10\%/10\%/80\% train/val/test data splits. Similarly, for the \texttt{lastfm-asia} \citep{rozemberczki2020feather} and \texttt{facebook} \citep{rozemberczki2019gemsec} datasets, as well as for our new \texttt{amazon-ratings-full} dataset, we use random stratified 10\%/10\%/80\% train/val/test data splits (see Appendix~\ref{app:amazon-ratings-full} for more details on the available \texttt{amazon-ratings-full} splits). The datasets \texttt{amazon-ratings-5core} and \texttt{questions} from \citet{platonov2023critical} originally come with 10 official random stratified 50\%/25\%/25\% train/val/test data splits. However, we notice that when, as in our work, extensive hyperparameter tuning on the validation set is performed (which is very important to achieve the best GNN performance in practice), using 10 different train/val/test splits leads to a certain type of test information leakage, as nodes that appear in the test data subset in one split can appear in validation subsets in other splits. Thus, instead of using 10 data splits, we use only one data split in our experiments. Specifically, we use the first data split from the official 10 data splits.

\subsection{Dataset characteristics}

The datasets used in our experiments are diverse in domains, sizes, and graph structural characteristics. Some structural characteristics of the considered graphs are provided in Table~\ref{tab:datasets-characteristics}. First, note that the datasets range in size from $7.6$K nodes to $1.6$M nodes and from $27.8$K edges to $22.3$M edges. We use both rather sparse and rather dense datasets, with average node degree ranging from $3.26$ to $460.92$ and median node degree ranging from $1$ to $71$. The average distance between two nodes in the considered graphs ranges from $2.45$ to $194.05$, while the maximum distance (graph diameter) ranges from $11$ to $553$. Clustering coefficients show how typical it is for two neighbors of a node to also be neighbors. There are two widely used definitions of clustering coefficients \citep{boccaletti2014structure}: the global clustering coefficient and the average local clustering coefficient. In graphs used in our experiments, both clustering coefficients range from approximately zero to rather high values. Note that high values of clustering coefficients are not required for the strong performance of CLATT, as it can bring substantial benefits even on datasets with approximately zero clustering coefficients such as \texttt{city-roads-M}, \texttt{city-roads-L}, and \texttt{questions}. The degree assortativity coefficient is the Pearson correlation coefficient of degrees for pairs of linked nodes. Among the considered datasets, there are graphs with both positive, approximately zero, and negative degree assortativity coefficients.

Let us also discuss the relationships between graph structure and node labels in the considered datasets. Datasets where nodes tend to connect to other nodes with similar labels are known as \emph{homophilous}, whereas datasets which do not exhibit this connectivity pattern are known as \emph{non-homophilous}. To measure the levels of homophily for regression datasets, we use the target assortativity coefficient, which is the Pearson correlation coefficient of targets for pairs of linked nodes. To measure the levels of homophily for classification datasets, we use unbiased homophily \citep{mironov2024revisiting} (with the $\alpha$ parameter set to $0$), which satisfies more properties desirable for a homophily measure from \citet{platonov2023characterizing} than other commonly used homophily measures. Among the datasets used in our experiments, there is a range of both homophilous datasets (\texttt{city-roads-M}, \texttt{city-roads-L}, \texttt{city-reviews}, \texttt{pokec-regions}, \texttt{lastfm-asia}, \texttt{facebook}) and non-homophilous datasets (\texttt{avazu-ctr}, \texttt{tolokers-2}, \texttt{hm-categories}, \texttt{questions}, \texttt{amazon-ratings-5core}, \texttt{amazon-ratings-full}). As our experimental results show, CLATT can substantially improve performance on both of these types of datasets.

Note that some of the graphs considered in our experiments are directed; however, we convert all graphs to undirected ones (by duplicating each edge with its direction reversed and then removing multiple edges) for clustering, training models, and also for reporting statistics in Table~\ref{tab:datasets-characteristics}.

\section{Amazon-ratings-full dataset}
\label{app:amazon-ratings-full}

The \texttt{amazon-ratings-5core} dataset from \citet{platonov2023critical} (originally called \texttt{amazon-ratings}) was obtained from data collected by \citet{leskovec2007dynamics}. Since this dataset was used by \citet{platonov2023critical} to evaluate models specifically designed for non-homophilous graphs, many of which are not scalable, only the 5-core of the graph \citep{malliaros2020core} was used, i.e., nodes of degree less than 5 were iteratively removed from the graph until no such nodes were left. While this procedure was used to reduce the size of the graph, we notice that it resulted in a graph with a peculiar structure: \texttt{amazon-ratings-5core} has a lot of small clusters of 5 or slightly more nodes that are very densely interconnected (often being cliques or almost cliques) but connected with the rest of the graph with only one or two edges. We hypothesize that such graph structure might be particularly amenable to node clustering and might overestimate the performance of CLATT (compared to its performance on the full co-purchasing network). Thus, we constructed a new dataset~--- \texttt{amazon-ratings-full}~--- which follows the same dataset construction process as used in \citet{platonov2023critical} but skips the reduction of the graph to its 5-core. Thus, we obtain the full Amazon co-purchasing network from \citet{leskovec2007dynamics} (more specifically, its full largest connected component, as the original data also contains many small connected components and isolated nodes) with the same task and node features as used by \citet{platonov2023critical}. As can be seen from Table~\ref{tab:datasets-characteristics}, \texttt{amazon-ratings-full} is more than an order of magnitude larger than \texttt{amazon-ratings-5core}, while having almost the same diameter but smaller average node degree, average distance between nodes, and unbiased homophily.

When creating data splits for our new dataset, we follow the approach from \citet{bazhenov2025graphland} and create 3 different data splits that can be used for different purposes. Specifically, we create 2 random stratified data splits with different proportions: the \texttt{RL} (random low) data split is a 10\%/10\%/80\% train/val/test data split, while the \texttt{RH} (random high) data split is a 50\%/25\%/25\% train/val/test data split. Further, the \texttt{TH} (temporal high) data split is a temporal data split with exactly the same proportions as the \texttt{RH} split, which can be used to evaluate model performance under the challenging setting of temporal distributional shifts. Since the original data does not have information about the time a product appeared on Amazon, we use the time the first review for a product appeared as a proxy for it to create the temporal data split. Additionally, the \texttt{THI} (temporal high / inductive) setting can be used with the \texttt{amazon-ratings-full} dataset by considering three snapshots of the graph taken at different timestamps for training, validation, and testing. In all our experiments in this work, we use the \texttt{RL} (random low) data split for the \texttt{amazon-ratings-full} dataset, similar to our setup for datasets from the GraphLand benchmark.

It can be seen that the improvements from CLATT are typically smaller on \texttt{amazon-ratings-full} than on \texttt{amazon-ratings-5core}, which suggests that indeed the specific graph structure of \texttt{amazon-ratings-5core} is particularly beneficial for CLATT. However, even on \texttt{amazon-ratings-full}, CLATT brings statistically significant increases in model performance for all the considered models.

\section{Experimental setup}
\label{app:experimental-setup}

For our experiments, we follow the experimental protocol from \citet{bazhenov2025graphland}. To compute the mean and standard deviation of model results, we train each model $10$ times with different random seeds, except for the largest considered dataset \texttt{pokec-regions}, for which we train each model 5 times.

We train all our models in a full-batch setting as is common for GNNs, i.e., we do not use any subgraph sampling methods and train the models on the full graph (in particular, for GraphSAGE we only use the model architecture but not the neighbor sampling technique).

Hyperparameter tuning is extremely important for achieving optimal performance with GNNs. Thus, we conduct a hyperparameter grid search on the validation set for all models. Specifically, for learning rate we consider the values $\{ 3 \times 10^{-5}, 1 \times 10^{-4}, 3 \times 10^{-4}, 1 \times 10^{-3}, 3 \times 10^{-3} \}$, and for dropout we consider the values of $\{ 0, 0.1, 0.2 \}$. For datasets from the GraphLand benchmark, we additionally consider transforming numerical features with either standard scaling or quantile transformation to standard normal distribution, and for regression datasets, we additionally consider either transforming targets with standard scaling or leaving them untransformed. For all models, we set the number of layers to $3$ and the hidden dimension to $512$. For GGTs, we set the dimension of positional encodings (DeepWalk embeddings or Laplacian eigenvectors) to $128$. We train all models for $1000$ steps with the Adam optimizer \citep{kingma2014adam}, except for the \texttt{amazon-ratings-5core} dataset, for which we have found longer training to often be beneficial, and thus train all models for $3000$ steps.

When using CLATT, for all clustering algorithms, we discard clusters that consist of less than 4 or more than 512 nodes, i.e., we do not apply CLATT to them.

Our model implementations use PyTorch \citep{paszke2019pytorch, ansel2024pytorch} and DGL \citep{wang2019deep}. All experiments were run on an NVIDIA Tesla A100 80GB GPU.

\section{Reproducibility}

We provide our code with the implementation of CLATT and instructions to reproduce all experimental results in our paper in the following repository: \url{https://github.com/OlegPlatonov/cluster-attention}.

\end{document}

%% file: inputs/tables/dataset_characteristics.tex
\resizebox{\textwidth}{!}{
\begin{tabular}{lrrrrrrrrrrrr}
& \multicolumn{3}{c}{node regression} & \multicolumn{9}{c}{node classification} \\
\cmidrule(lr){2-4}
\cmidrule(lr){5-13}
& \rotatebox{90}{\footnotesize city-roads-M} & 
\rotatebox{90}{\footnotesize city-roads-L} & 
\rotatebox{90}{\footnotesize avazu-ctr} &
\rotatebox{90}{\footnotesize tolokers-2} & 
\rotatebox{90}{\footnotesize city-reviews} & 
\rotatebox{90}{\footnotesize hm-categories} & 
\rotatebox{90}{\footnotesize pokec-regions} & 
\rotatebox{90}{\footnotesize questions} & 
\rotatebox{90}{\footnotesize lastfm-asia} & 
\rotatebox{90}{\footnotesize facebook} &
\rotatebox{90}{\footnotesize amzn-r-5core} &
\rotatebox{90}{\footnotesize amzn-r-full} \\
\midrule
\# nodes & $57.1$K & $142.3$K & $76.3$K & $11.8$K & $148.8$K & $46.5$K & $1.6$M & $48.9$K & $7.6$K & $22.5$K & $24.5$K & $334.9$K \\
\# edges & $107.1$K & $231.6$K & $11.0$M & $519.0$K & $1.2$M & $10.7$M & $22.3$M & $153.5$K & $27.8$K & $170.8$K & $93.1$K & $925.8$K \\
avg degree & $3.75$ & $3.26$ & $288.04$ & $88.28$ & $15.66$ & $460.92$ & $27.32$ & $6.28$ & $7.29$ & $15.20$ & $7.60$ & $5.53$ \\
median degree & $4$ & $3$ & $71$ & $30$ & $4$ & $45$ & $13$ & $1$ & $4$ & $7$ & $5$ & $4$ \\
avg distance & $126.75$ & $194.05$ & $3.55$ & $2.79$ & $4.91$ & $2.45$ & $4.68$ & $4.29$ & $5.23$ & $4.97$ & $16.24$ & $11.95$ \\
diameter & $383$ & $553$ & $14$ & $11$ & $19$ & $13$ & $14$ & $16$ & $15$ & $15$ & $46$ & $47$ \\
global clustering & $0.00$ & $0.00$ & $0.24$ & $0.23$ & $0.26$ & $0.27$ & $0.05$ & $0.02$ & $0.18$ & $0.23$ & $0.32$ & $0.21$ \\
avg local clustering & $0.00$ & $0.00$ & $0.85$ & $0.53$ & $0.41$ & $0.70$ & $0.11$ & $0.03$ & $0.22$ & $0.36$ & $0.58$ & $0.40$ \\
degree assortativity & $0.70$ & $0.74$ & $-0.30$ & $-0.08$ & $0.01$ & $-0.35$ & $0.00$ & $-0.15$ & $0.02$ & $0.08$ & $-0.09$ & $-0.06$ \\
\midrule
\# classes & N/A & N/A & N/A & $2$ & $2$ & $21$ & $183$ & $2$ & $18$ & $4$ & $5$ & $5$ \\
unbiased homophily & N/A & N/A & N/A & $0.10$ & $0.69$ & $0.38$ & $0.98$ & $0.06$ & $0.97$ & $0.90$ & $0.28$ & $0.22$ \\
target assortativity & $0.74$ & $0.72$ & $0.18$ & N/A & N/A & N/A & N/A & N/A & N/A & N/A & N/A & N/A \\
\bottomrule
\end{tabular}
}

%% file: inputs/tables/results_graphland.tex
\resizebox{\textwidth}{!}{
\begin{tabular}{lccccccc}
& \multicolumn{3}{c}{regression} & \multicolumn{2}{c}{bin. class.} & \multicolumn{2}{c}{mult. class.} \\
\cmidrule(lr){2-4}
\cmidrule(lr){5-6}
\cmidrule(lr){7-8}
& city-roads-M & city-roads-L & avazu-ctr & tolokers-2 & city-reviews & hm-categories & pokec-regions \\
\midrule
GCN & $59.05 \pm 0.16$ & $53.26 \pm 0.14$ & $30.47 \pm 0.27$ & $51.48 \pm 0.81$ & $77.15 \pm 0.28$ & $61.88 \pm 0.23$ & $34.99 \pm 0.17$ \\
GCN-CLATT & $\underline{\mathbf{59.78 \pm 0.28}}$ & $\underline{\mathbf{55.15 \pm 0.20}}$ & $\mathbf{30.74 \pm 0.26}$ & $\underline{\mathbf{54.08 \pm 0.61}}$ & $\underline{\mathbf{77.79 \pm 0.18}}$ & $\underline{\mathbf{64.96 \pm 0.24}}$ & $\underline{\mathbf{48.88 \pm 0.44}}$ \\
\\
GraphSAGE & $57.51 \pm 0.53$ & $52.43 \pm 0.25$ & $31.84 \pm 0.24$ & $53.87 \pm 0.78$ & $77.82 \pm 0.13$ & $56.72 \pm 0.31$ & $37.80 \pm 0.35$ \\
GraphSAGE-CLATT & $\underline{\mathbf{60.67 \pm 0.18}}$ & $\underline{\mathbf{55.59 \pm 0.24}}$ & $\underline{\mathbf{32.44 \pm 0.22}}$ & $\mathbf{54.76 \pm 0.63}$ & $\mathbf{77.91 \pm 0.12}$ & $\underline{\mathbf{61.90 \pm 0.25}}$ & $\underline{\mathbf{48.75 \pm 0.91}}$ \\
\\
LGT & $58.05 \pm 0.58$ & $53.38 \pm 0.12$ & $30.87 \pm 0.47$ & $55.70 \pm 0.28$ & $76.97 \pm 0.21$ & $69.25 \pm 0.25$ & $46.41 \pm 0.20$ \\
LGT-CLATT & $\underline{\mathbf{60.26 \pm 0.28}}$ & $\underline{\mathbf{55.75 \pm 0.34}}$ & $\mathbf{31.48 \pm 0.45}$ & $\underline{\mathbf{56.95 \pm 0.28}}$ & $\underline{\mathbf{77.40 \pm 0.16}}$ & $\underline{\mathbf{70.22 \pm 0.29}}$ & $\underline{\mathbf{49.34 \pm 0.29}}$ \\
\\
GGT-DW & $53.14 \pm 0.50$ & $48.24 \pm 0.82$ & $27.78 \pm 0.62$ & $56.27 \pm 0.34$ & $74.70 \pm 0.19$ & $41.72 \pm 0.41$ & TL \\
GGT-DW-CLATT & $\underline{\mathbf{54.58 \pm 0.44}}$ & $\mathbf{48.72 \pm 0.41}$ & $\mathbf{28.71 \pm 0.41}$ & $\underline{\mathbf{57.93 \pm 0.45}}$ & $\underline{\mathbf{76.25 \pm 0.20}}$ & $\underline{\mathbf{55.44 \pm 0.34}}$ & TL \\
\\
GGT-Lap & $54.32 \pm 0.64$ & $46.48 \pm 0.99$ & $24.27 \pm 1.16$ & $49.46 \pm 1.05$ & $71.40 \pm 0.36$ & $37.79 \pm 0.34$ & TL \\
GGT-Lap-CLATT & $\underline{\mathbf{57.53 \pm 0.18}}$ & $\underline{\mathbf{49.86 \pm 0.34}}$ & $\underline{\mathbf{25.98 \pm 0.33}}$ & $\underline{\mathbf{53.77 \pm 0.82}}$ & $\underline{\mathbf{76.62 \pm 0.16}}$ & $\underline{\mathbf{52.80 \pm 0.40}}$ & TL \\
\bottomrule
\end{tabular}
}

%% file: inputs/tables/results_other.tex
\resizebox{0.85\textwidth}{!}{
\begin{tabular}{lccccc}
& \multicolumn{1}{c}{bin. class.} & \multicolumn{4}{c}{mult. class.} \\
\cmidrule(lr){2-2}
\cmidrule(lr){3-6}
& questions & lastfm-asia & facebook & amazon-ratings-5core & amazon-ratings-full \\
\midrule
GCN & $18.60 \pm 0.59$ & $80.62 \pm 0.36$ & $91.00 \pm 0.14$ & $49.65 \pm 0.50$ & $39.28 \pm 0.13$ \\
GCN-CLATT & $\underline{\mathbf{20.48 \pm 0.83}}$ & $\underline{\mathbf{84.01 \pm 0.32}}$ & $\underline{\mathbf{91.71 \pm 0.21}}$ & $\underline{\mathbf{52.84 \pm 0.53}}$ & $\underline{\mathbf{39.76 \pm 0.13}}$ \\
\\
GraphSAGE & $20.41 \pm 0.47$ & $80.90 \pm 0.46$ & $91.22 \pm 0.21$ & $\mathbf{54.94 \pm 0.38}$ & $38.73 \pm 0.12$ \\
GraphSAGE-CLATT & $\mathbf{20.89 \pm 0.30}$ & $\underline{\mathbf{84.54 \pm 0.19}}$ & $\underline{\mathbf{92.09 \pm 0.24}}$ & $54.70 \pm 0.43$ & $\underline{\mathbf{39.21 \pm 0.11}}$ \\
\\
LGT & $17.86 \pm 0.52$ & $83.97 \pm 0.27$ & $92.17 \pm 0.20$ & $51.26 \pm 0.38$ & $39.03 \pm 0.08$ \\
LGT-CLATT & $\underline{\mathbf{21.28 \pm 0.68}}$ & $\underline{\mathbf{84.55 \pm 0.21}}$ & $\underline{\mathbf{92.71 \pm 0.13}}$ & $\underline{\mathbf{53.70 \pm 0.49}}$ & $\underline{\mathbf{39.34 \pm 0.17}}$ \\
\\
GGT-DW & $16.26 \pm 0.72$ & $78.69 \pm 0.50$ & $86.84 \pm 0.27$ & $46.22 \pm 0.79$ & TL \\
GGT-DW-CLATT & $\underline{\mathbf{19.69 \pm 1.09}}$ & $\underline{\mathbf{83.31 \pm 0.42}}$ & $\underline{\mathbf{91.31 \pm 0.27}}$ & $\underline{\mathbf{48.42 \pm 0.37}}$ & TL \\
\\
GGT-Lap & $18.38 \pm 0.71$ & $63.50 \pm 1.16$ & $73.73 \pm 0.73$ & $45.97 \pm 0.71$ & TL \\
GGT-Lap-CLATT & $\underline{\mathbf{21.20 \pm 0.51}}$ & $\underline{\mathbf{82.43 \pm 0.37}}$ & $\underline{\mathbf{89.90 \pm 0.22}}$ & $\underline{\mathbf{52.03 \pm 0.50}}$ & TL \\
\bottomrule
\end{tabular}
}

%% file: inputs/tables/attention.tex
\resizebox{\textwidth}{!}{
\begin{tabular}{lcccccccccc}
& \multicolumn{5}{c}{lastfm-asia} & \multicolumn{5}{c}{tolokers-2} \\
\cmidrule(lr){2-6}
\cmidrule(lr){7-11}
\multicolumn{1}{r}{\textit{quantile} $\rightarrow$} & $0.05$ & $0.25$ & $0.50$ & $0.75$ & $0.95$ & $0.05$ & $0.25$ & $0.50$ & $0.75$ & $0.95$ \\
\midrule
LGT (local attention) & $0.39$ & $0.62$ & $0.79$ & $0.90$ & $0.97$ & $0.59$ & $0.91$ & $0.97$ & $0.99$ & $1.00$ \\
LGT-CLATT (local attention) & $0.38$ & $0.60$ & $0.78$ & $0.89$ & $0.96$ & $0.56$ & $0.91$ & $0.98$ & $0.99$ & $1.00$ \\
LGT-CLATT (cluster attention) & $1.35$ & $1.99$ & $2.56$ & $3.50$ & $5.23$ & $1.23$ & $1.73$ & $1.99$ & $2.32$ & $3.14$ \\
GGT-DW (global attention) & $4.13$ & $4.71$ & $5.16$ & $5.67$ & $6.61$ & $2.23$ & $2.49$ & $2.71$ & $2.93$ & $3.35$ \\
GGT-DW-CLATT (global attention) & $4.10$ & $4.69$ & $5.13$ & $5.65$ & $6.60$ & $2.26$ & $2.52$ & $2.73$ & $2.94$ & $3.36$ \\
GGT-DW-CLATT (cluster attention) & $1.39$ & $2.01$ & $2.60$ & $3.55$ & $5.25$ & $1.23$ & $1.73$ & $2.00$ & $2.33$ & $3.16$ \\
GGT-Lap (global attention) & $4.18$ & $4.71$ & $5.13$ & $5.65$ & $6.59$ & $1.96$ & $2.15$ & $2.82$ & $3.00$ & $3.90$ \\
GGT-Lap-CLATT (global attention) & $4.16$ & $4.72$ & $5.14$ & $5.67$ & $6.62$ & $2.24$ & $2.50$ & $2.71$ & $2.93$ & $3.34$ \\
GGT-Lap-CLATT (cluster attention) & $1.19$ & $1.86$ & $2.32$ & $3.07$ & $4.70$ & $1.20$ & $1.72$ & $1.99$ & $2.32$ & $3.14$ \\
\bottomrule
\end{tabular}
}

%% file: inputs/tables/clustering_algorithms.tex
\begin{tabular}{lcccc}
\multicolumn{1}{r}{\textit{clustering algorithm} $\rightarrow$} & LA & BPP & H1 & KM \\
\midrule
city-roads-M & \textcolor{green}{\checkmark} & \textcolor{green}{\checkmark} &  &  \\
city-roads-L & \textcolor{green}{\checkmark} & \textcolor{green}{\checkmark} &  &  \\
avazu-ctr & \textcolor{green}{\checkmark} &  &  &  \\
tolokers-2 &  & \textcolor{green}{\checkmark} &  &  \\
city-reviews &  &  & \textcolor{green}{\checkmark} &  \\
hm-categories & \textcolor{green}{\checkmark} &  &  &  \\
pokec-regions &  &  & \textcolor{green}{\checkmark} &  \\
questions & \textcolor{green}{\checkmark} &  &  & \textcolor{green}{\checkmark} \\
lastfm-asia & \textcolor{green}{\checkmark} & \textcolor{green}{\checkmark} & \textcolor{green}{\checkmark} &  \\
facebook & \textcolor{green}{\checkmark} &  & \textcolor{green}{\checkmark} &  \\
amazon-ratings-5core & \textcolor{green}{\checkmark} & \textcolor{green}{\checkmark} &  &  \\
amazon-ratings-full & \textcolor{green}{\checkmark} &  &  &  \\
\bottomrule
\end{tabular}